\documentclass{article}

\usepackage{arxiv}

\usepackage{amsmath}
\usepackage{amssymb}
\usepackage{multicol}
\usepackage{multirow}
\usepackage{booktabs}
\usepackage{enumitem}

\usepackage[utf8]{inputenc} % allow utf-8 input
\usepackage[T1]{fontenc}    % use 8-bit T1 fonts
\usepackage{hyperref}       % hyperlinks
\usepackage{url}            % simple URL typesetting
\usepackage{booktabs}       % professional-quality tables
\usepackage{amsfonts}       % blackboard math symbols
\usepackage{nicefrac}       % compact symbols for 1/2, etc.
\usepackage{microtype}      % microtypography
\usepackage{cleveref}       % smart cross-referencing
\usepackage{lipsum}         % Can be removed after putting your text content
\usepackage{graphicx}
\usepackage{natbib}
\usepackage{doi}

\graphicspath{ {./figures} }

\title{TextMastero: \\ Mastering High-Quality Scene Text Editing \\ in Diverse Languages and Styles}

\date{}

\newif\ifuniqueAffiliation
\ifuniqueAffiliation % Standard variant of author block
\author{
}
\else
\usepackage{authblk}

\author[]{%
	\hspace{1mm}Tong Wang%
}
\author[]{%
	\hspace{1mm}Xiaochao Qu%
}
\author[]{%
	\hspace{1mm}Ting Liu%
}
\vspace{2mm}
\affil[]{MT Lab}
\fi

\renewcommand{\headeright}{Preprint}
\renewcommand{\undertitle}{}
\renewcommand{\shorttitle} % {\textit{arXiv} Template}

\hypersetup{
pdftitle={TextMastero: Mastering High-Quality Scene Text Editing in Diverse Languages and Styles},
pdfsubject={cs.CV},
pdfauthor={Tong Wang},
pdfkeywords={Scene Text Editing}
}

\begin{document}
\maketitle

\begin{abstract}
Scene text editing aims to modify texts on images while maintaining the style of newly generated text similar to the original. Given an image, a target area, and target text, the task produces an output image with the target text in the selected area, replacing the original.  This task has been studied extensively, with initial success using Generative Adversarial Networks (GANs) to balance text fidelity and style similarity.  However, GAN-based methods struggled with complex backgrounds or text styles. Recent works leverage diffusion models, showing improved results, yet still face challenges, especially with non-Latin languages like CJK characters (Chinese, Japanese, Korean) that have complex glyphs, often producing inaccurate or unrecognizable characters. To address these issues, we present \emph{TextMastero} - a carefully designed multilingual scene text editing architecture based on latent diffusion models (LDMs). TextMastero introduces two key modules: a glyph conditioning module for fine-grained content control in generating accurate texts, and a latent guidance module for providing comprehensive style information to ensure similarity before and after editing. Both qualitative and quantitative experiments demonstrate that our method surpasses all known existing works in text fidelity and style similarity.
\end{abstract}

%%%%%% 
\section{Introduction}

\begin{figure}[h]
\centering
\includegraphics[width=0.9\columnwidth]{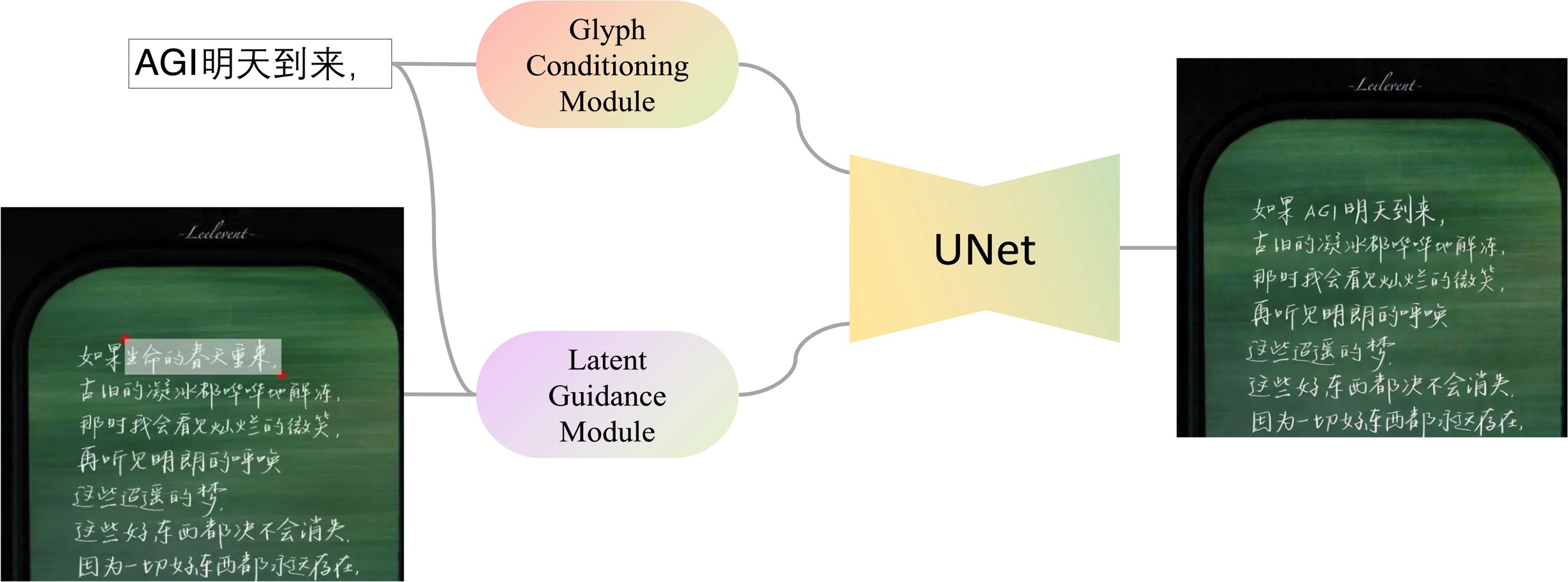} % Reduce the figure size so that it is slightly narrower than the column. Don't use precise values for figure width.This setup will avoid overfull boxes.
\caption{Overall Architecture}
\label{fig1}
\end{figure}

Scene text editing, a sub-field of general image editing, is a challenging task that aims to modify text in images while preserving the original style and maintaining visual coherence. It has significant applications for both regular users and professional designers, especially for those typefaces that are hard to find or even unable to be defined. Despite recent advancements, achieving high-quality results across diverse languages and complex visual scenarios remains a formidable challenge, particularly for non-Latin scripts with intricate glyph structures.

Early successful works such as ~\cite{DBLP:journals/corr/abs-2107-11041}, ~\cite{DBLP:conf/mm/WuZLHLDB19}, ~\cite{DBLP:conf/cvpr/RoyBG020}, ~\cite{DBLP:conf/aaai/QuTXXW023} utilized generative adversarial networks (GANs) to accomplish the task and obtained some good results. They generally formulate the problem as a style transfer problem. Concretely, given an image and a region to edit, the cropped region from the image is referred to as the reference image. The training of the GAN can be summarized as training the renderer, or generator, $G$, to write text that can fool the discriminator, $D$, so that the generated image would contain new text content while maintaining a similar style to the reference image. However, due to style complexities such as typefaces, color, affine and non-affine transformations, etc., and the limitations of GAN's generation ability, these methods fail on complex editing scenarios.

Diffusion models ~\cite{DBLP:conf/nips/HoJA20} are another paradigm for image generation. Compared to GANs, images produced with this paradigm generally have better quality, partly because of their iterative denoising nature. Recent works that utilized diffusion models such as ~\cite{DBLP:journals/corr/abs-2304-05568}, ~\cite{DBLP:conf/nips/ChenXGLZLMZW23}, ~\cite{DBLP:conf/iclr/TuoXHGX24} have made extraordinary improvements in both text fidelity and style coherency in the scene text editing task. However, they still fail on complicated scenes due to reasons described earlier. Moreover, they are unable to generate small characters and long text contents. Most importantly, for non-Latin characters like CJK Characters, they produce unsatisfactory results too frequently because of the intricate glyph structures, showcased by wrongly written characters (similar to misspelled words in Latin languages but in two dimensions).

% %%%
\begin{figure*}[t]
    \centering
    \includegraphics[width=1\textwidth]{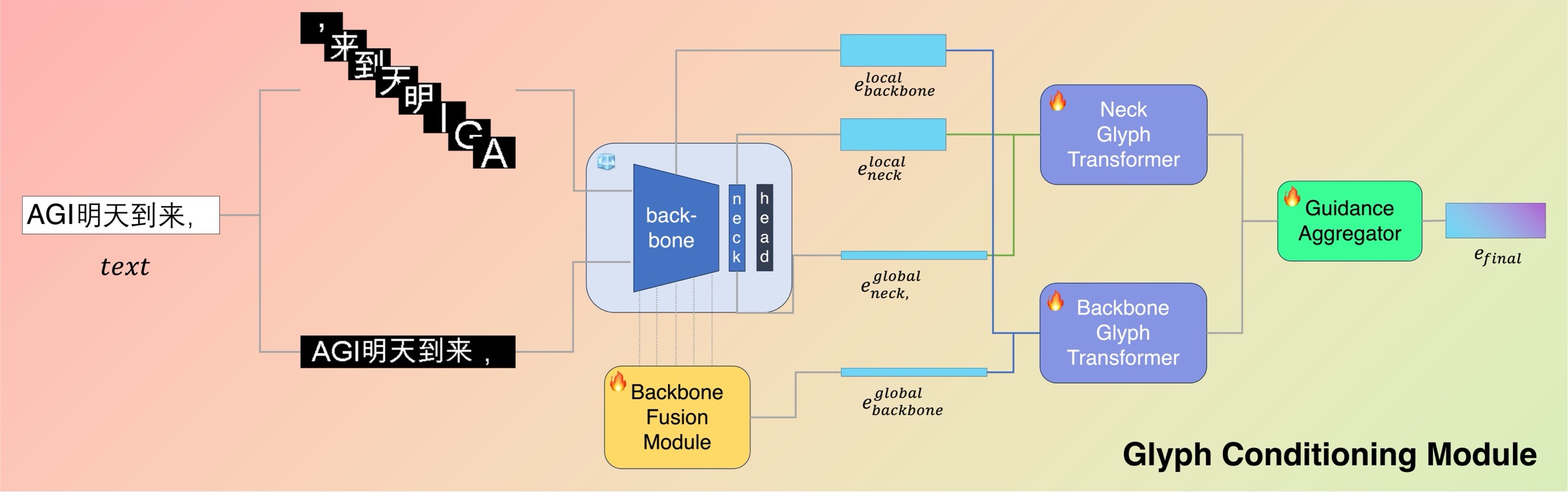}
    % \caption{Left: Glyph Conditioning Module. Right: Latent Guidance Module.}
    % \hfill
    \includegraphics[width=1\textwidth]{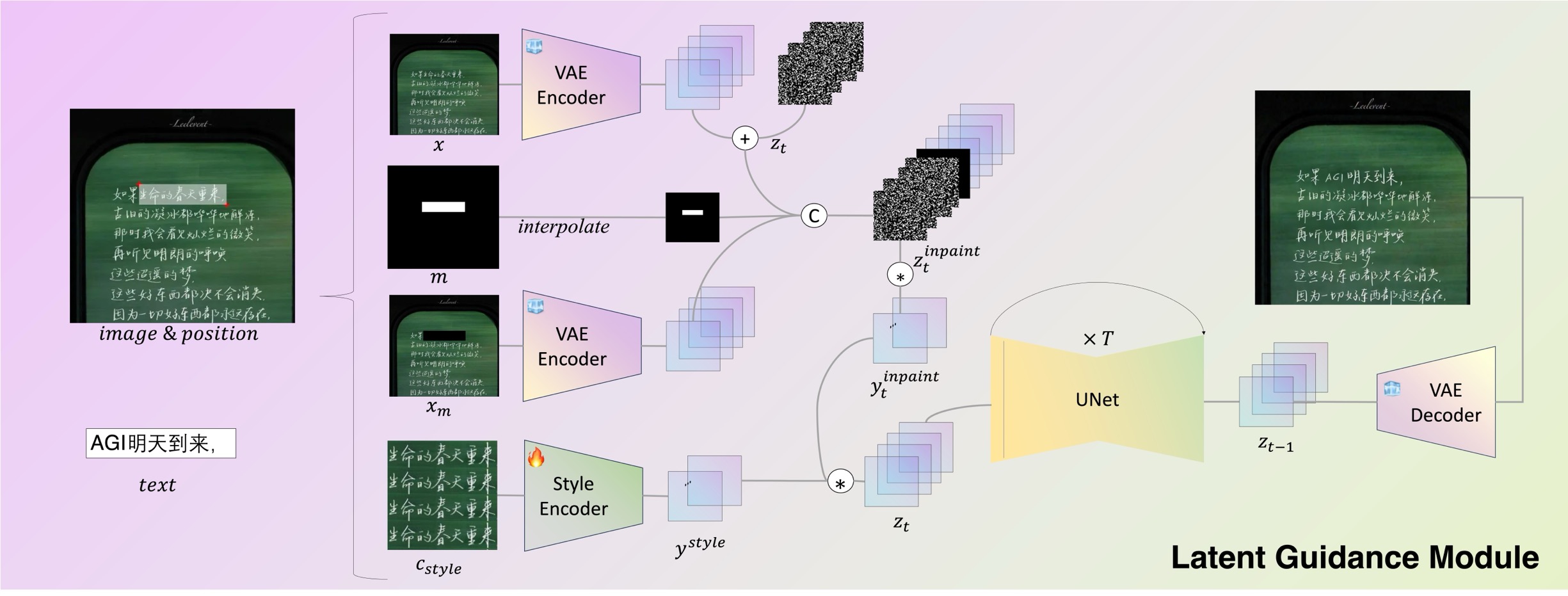}
    \caption{Complete model architecture of \emph{Textmastero}. Top: The glyph conditioning module enables fine-grained content control. Bottom: The latent guidance module maintains style consistency}
    \label{fig:combined}
\end{figure*}

In this work, we propose a novel framework \emph{TextMastero}, producing state-of-the-art results in the multilingual scene text editing task. Both quantitative and qualitative comparisons show that our method is superior to previous state-of-the-art methods. The main innovations of our \emph{TextMastero} framework are as follows:
\begin{itemize}
\item We introduce character-line modeling for the input text, then use the interaction between them as the condition for the generation model for character-level content control
\item We utilize multilevel OCR model's backbone features for glyph-level fine-grained control
\item We propose a style encoder to encode the text region as additional style guidance for the UNet for better style coherency
\end{itemize}
Our contributions can be summarized into two modules: Glyph Conditioning Module and Latent Guidance Module (see Section 3 for details). Ablation studies has also shown effectiveness and necessity for each of our design.

\begin{figure*}[t]
\centering
\includegraphics[width=1\textwidth]{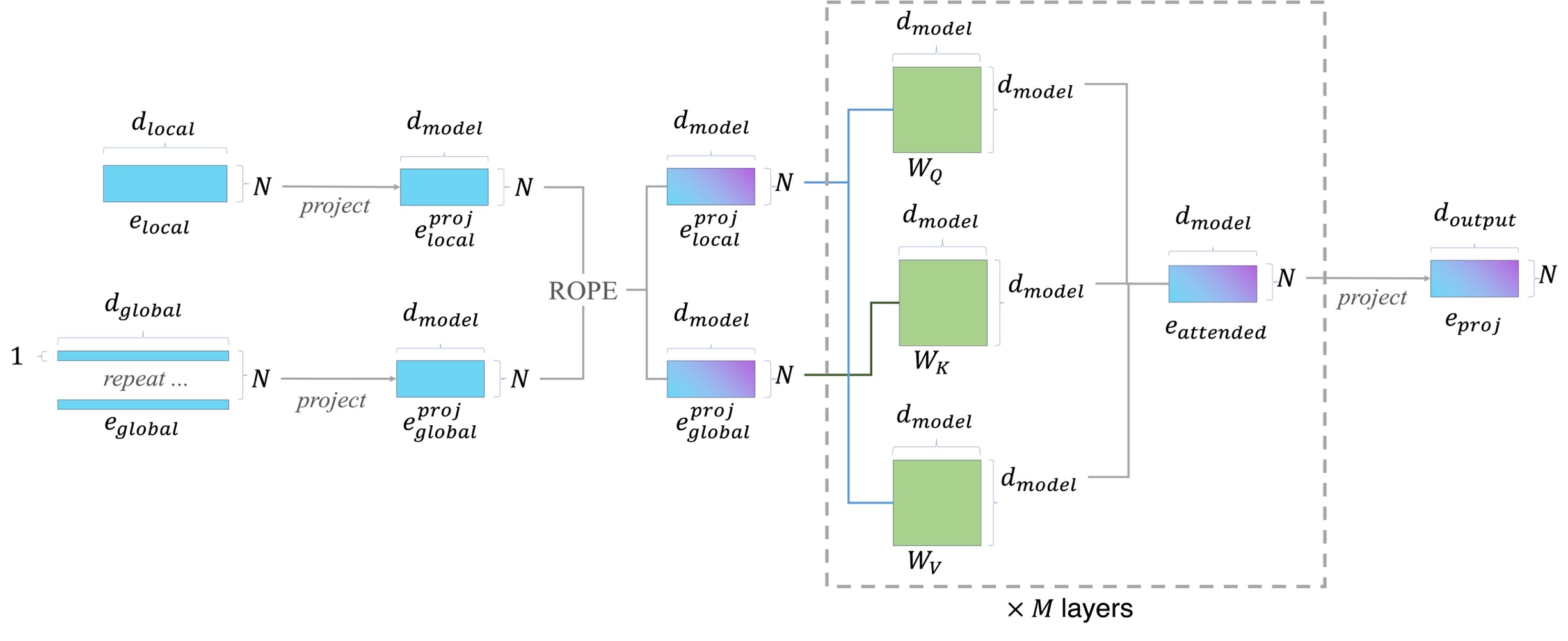} 
\caption{Glyph Transformer}
\label{fig3}
\end{figure*}

\section{Related Work}
\subsection{Scene Text Editing}
Scene text editing has been approached through two main paradigms: GAN-based ~\cite{DBLP:journals/corr/abs-2107-11041}, ~\cite{DBLP:conf/mm/WuZLHLDB19}, ~\cite{DBLP:conf/cvpr/RoyBG020}, ~\cite{DBLP:conf/aaai/QuTXXW023} and diffusion-based ~\cite{DBLP:journals/corr/abs-2304-05568}, ~\cite{DBLP:conf/nips/ChenXGLZLMZW23}, ~\cite{DBLP:conf/iclr/TuoXHGX24}. GAN-based works, such as STEFANN ~\cite{DBLP:conf/cvpr/RoyBG020}, utilize style transfer concepts by employing a font-adaptive model to preserve text structures and another to extract and transfer style information.

Diffusion-based models employ two primary conditioning mechanisms: cross-attention guidance and direct latent space guidance. For cross-attention, AnyText~\cite{DBLP:journals/corr/abs-2311-03054} uses a pretrained OCR model to extract features from a rendered glyph image, incorporating these alongside image descriptions encoded by CLIP model's text branch. DiffUTE replaces CLIP guidance with fixed-length features from TrOCR's ~\cite{DBLP:conf/aaai/LiLC0LFZ0W23} last hidden state. DiffSTE \cite{DBLP:journals/corr/abs-2304-05568} proposes a dual cross-attention mechanism using CLIP embeddings and a character encoder. Our work utilizes the same pretrained OCR model as AnyText but introduces character-level fine-grained features with various techniques (detailed in section 3.1).

Direct latent space guidance, exemplified by ControlNet ~\cite{DBLP:conf/iccv/ZhangRA23}, encodes input conditions via convolution layers and adds them to original latents. However, this approach tends to align the generated image with the condition's geometry, which is undesirable for scene text editing as we only want to encode style of the text region, not geometry contents in it. Thus, following ~\cite{DBLP:journals/corr/abs-2311-03054}, we provide latent conditions in an indirect way by concatenating latent style features to original UNet's input convolution feature map, followed by another convolution layer to project back to UNet's original dimension.

\subsection{Diffusion Models}
Diffusion models in deep learning were inspired by non-equilibrium thermodynamics. The forward pass destroys data by adding noise iteratively with some schedule, while the backward pass, which is the learning process, tries to restore the data. After training, we get a generative model since we can start from noise to generate new data samples. DDPM ~\cite{DBLP:conf/nips/HoJA20} is the first work that used diffusion models for image generation and has become the foundation of most of today's diffusion models. Since its introduction, diffusion models for image generation have diverged into two directions: pixel space denoising models such as Imagen ~\cite{DBLP:conf/nips/SahariaCSLWDGLA22}, and latent space models (LDMs) ~\cite{DBLP:conf/cvpr/RombachBLEO22} that firstly encode images to latent space, denoise the latents, then decode them back to image space. The latter is less computationally intensive, hence it is more popular than the former for downstream tasks. Our work is based on a pretrained text-to-image LDM model. The training objective of LDMs can be written as:
$$L_{\text{LDM}} := \mathbb{E}_{\mathcal{E}(x), c, \epsilon \sim \mathcal{N}(0,1), t} \left[ \left\| \epsilon - \epsilon_\theta(z_t, c, t) \right\|_2^2 \right] .
$$
$\mathcal{E}$ is the encoder that compresses $x$ to get latents $z$. $\epsilon$ represents the ground truth noise that was added to $z$ in the forward pass. $\epsilon_\theta$ is the time-dependent model that predicts the noise added given time step $t$, realized by a UNet~\cite{DBLP:conf/miccai/RonnebergerFB15}. $c$ is the extra condition that guides the denoising process. Once the denoising process is finished, we decode the latents with $\hat{x} = \mathcal{D}(\hat{z})$ to get the resulting image. Our work can be summarized as designing novel glyph conditions $c$ and additional style guidance that operates on latents $z_t$.

\subsection{Feature Fusion}
Feature fusion is a common technique in computer vision to combine complementary information from different sources or representations. Early approaches like skip connections ~\cite{DBLP:conf/cvpr/HeZRS16} and Feature Pyramid Networks ~\cite{DBLP:conf/cvpr/LinDGHHB17} demonstrated significant performance improvements by combining features between different layers and scales of the network. Recent works stress the same thing. ~\cite{jiang2024clipdinovisualencoders} analyzed feature maps of CLIP ~\cite{DBLP:conf/icml/RadfordKHRGASAM21} and DINOv2 ~\cite{DBLP:journals/corr/abs-2304-07193} in detail and found that the intermediate representations of different layers have distinct characteristics. For example, attention maps of CLIP show that shallow layers focus on local features while the deep layers focus on global ones. This suggests that not only is feature fusion useful in convolutional networks, but also in transformer-based vision models, encompassing most commonly used neural networks today. Thus, we propose two types of feature fusion to fully utilize the OCR recognition model's capability. We fuse backbone features with a FPN-like fusion module and combine backbone and neck features with our proposed Glyph Transformer. See section 3.1 for details.

\section{Method}

Figure \ref{fig1} presents the overall architecture of our proposed approach, which builds upon the latent diffusion models introduced by~\cite{DBLP:conf/cvpr/RombachBLEO22}. For clarity, we have omitted the VAE in this figure, but it is included in Figure \ref{fig:combined}. Our framework introduces two key modules tailored for scene text editing: a glyph guidance module that conditions text content, and a latent guidance module that handles text position and style conditions. These modules work together to enhance the model's ability to edit scene text while preserving the desired content, position, and style characteristics.

\subsection{Glyph Conditioning Module}
Unlike text-to-image LDMs that primarily use CLIP~\cite{DBLP:conf/icml/RadfordKHRGASAM21} for conditioning, we remove CLIP guidance as it's not ideal for scene text editing. Instead, following ~\cite{DBLP:journals/corr/abs-2311-03054}, we incorporate the pretrained PaddleOCR-v4 recognition model ~\cite{paddleocr2023} for encoding input texts. We introduce more fine-grained controls by leveraging the OCR model's capabilities. As shown in Figure \ref{fig:combined} (up), given a text input $x$, we render it to $x_{\text{local}} \in \mathbb{R}^{N \times 36 \times 48}$, a series of single-character glyph images, and $x_{\text{global}} \in \mathbb{R}^{48 \times 320}$, a full text line glyph image. These are fed into the OCR model, yielding backbone and neck features:
$$F_{OCR}(x_{local}) = (e^{local}_{backbone}, e^{local}_{neck})$$
$$F_{OCR}(x_{global}) = (e^{global}_{backbone}, e^{global}_{neck})$$
We then use two glyph transformers to capture interactions between local and global features for both backbone and neck modules. Finally, an aggregator $A$ projects and concatenates backbone $e_{backbone} \in \mathbb{R}^{N \times 1024}$ and neck $e_{neck} \in \mathbb{R}^{N \times 1024}$ features to obtain $e_{final} \in \mathbb{R}^{N \times 1024}$ as the cross-attention guide for the UNet.

\subsubsection{Glyph Transformer}
The intention of designing the Glyph Transformer is that we can use cross-attention to capture the interaction between character-level local features and line-level global features to get a better representation of text glyphs. This is inspired by Cutie~\cite{DBLP:journals/corr/abs-2310-12982}, a video segmentation work that used a cross-attention mechanism to capture relationships between local features at the pixel level and global features at the object level and eventually achieved state-of-the-art results in various video segmentation metrics and test datasets.

Figure \ref{fig3} illustrates the details of our Glyph Transformer. Given the $d_{local}$ and $d_{global}$ features extracted by $F_{OCR}$, we first repeat $1 \times d_{global}$ to $N \times d_{global}$ to match the $N \times d_{local}$ local features. Then we project both of them to the $d_{model}$ size. Next, we transform these projected features through learned linear projections where local features generate queries (Q) and values (V), while global features produce keys (K). We then apply rotary positional embedding (ROPE) to Q and K, effectively computing:
$Q', K' = \text{ROPE}(Q, K)$
where $Q'$ and $K'$ are the positionally-encoded versions of Q and K, respectively.

Experimentally, we found that without adding positional embedding, the generated text would have been disordered, meaning that characters would appear in random order. The choice of using ROPE instead of other positional embedding techniques is due to the fact that it integrates positional information after projection, avoiding the need to add embeddings twice for differently dimensioned local and global features.
We then perform standard cross-attention. This mechanism enables the model to leverage global context to enhance the understanding of local features in relation to text glyphs. The resulting attended features, $e_{attended}$, are in the $d_{model}$ dimension. Finally, we project these features to the desired output dimension $d_{output}$ to obtain the final conditions.

\subsubsection{Backbone Fusion Module}
To produce enhanced $e^{global}$ features, we designed a feature fusion module for the backbone, which is inspired by feature pyramid networks. This module can be generalized to various pretrained OCR recognition models, with our implementation specifically tailored for the PaddleOCR-V4 model.

In general, given a set of multiscale features ${x_1, x_2, ..., x_N}$ extracted from a backbone model, where $x_i \in \mathbb{R}^{C_i \times H_i \times W_i}$, our fusion module first projects each feature map to a common dimension $D$: $c_i = f_i(x_i)$, where $f_i: \mathbb{R}^{C_i \times H_i \times W_i} \rightarrow \mathbb{R}^{D \times H_i \times W_i}$. The features are then fused in a top-down manner, starting with $p_N = c_N$, and for $i=N-1, ..., 1$: $p_i = g_i(u(p_{i+1}) + c_i)$. Finally, the output is generated as $y = h(d(k(p_1)))$. Here, $u(\cdot)$ represents upsampling, $g_i$ are fusion operations, $k$ is a final projection, $d$ is downsampling, and $h$ is a pooling operation.

For our specific implementation with the PaddleOCR-V4 model, we use five levels of features ${x_1, x_2, x_3, x_4, x_5}$, where $x_i \in \mathbb{R}^{C_i \times H_i \times W_i}$. We apply 1x1 convolutions for the lateral connections: $c_i = f_i(x_i)$, where $f_i: \mathbb{R}^{C_i \times H_i \times W_i} \rightarrow \mathbb{R}^{D \times H_i \times W_i}$. Starting from the top level, we have $p_5 = c_5$, and for $i = 4, 3, 2, 1$: $p_i = g_i(u(p_{i+1}) + c_i)$, where $g_i$ is a 3x3 convolution. The final output is obtained by $y = h(d(k(p_1)))$, where $k$ is a 1x1 convolution, $d$ is downsampling, and $h$ is adaptive average pooling.

This model design adapts easily to various pretrained OCR recognition models while improving the quality of $e^{global}$ features. Our Backbone Fusion Module leverages multi-level feature maps to capture fine-grained representations of text characters, potentially enhancing text recognition accuracy. In our ablation studies, we observed clear improvements in the model's text generation performance, demonstrating the module's effectiveness.

\begin{figure}[htbp]
\centering
\begin{tabular}{@{}c@{\hspace{1mm}}c@{\hspace{1mm}}c@{\hspace{1mm}}c@{}}
% First row
\includegraphics[width=0.11\textwidth]{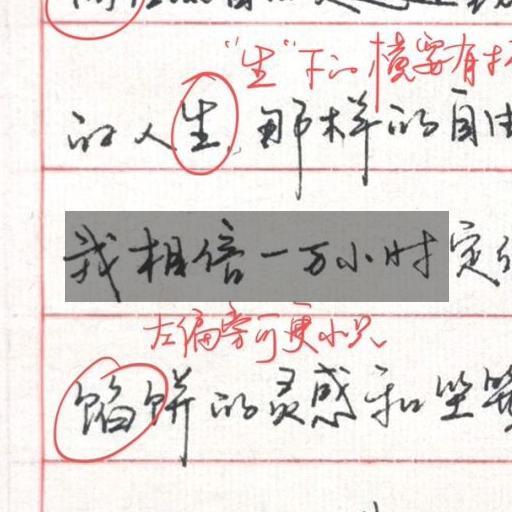} &
\includegraphics[width=0.11\textwidth]{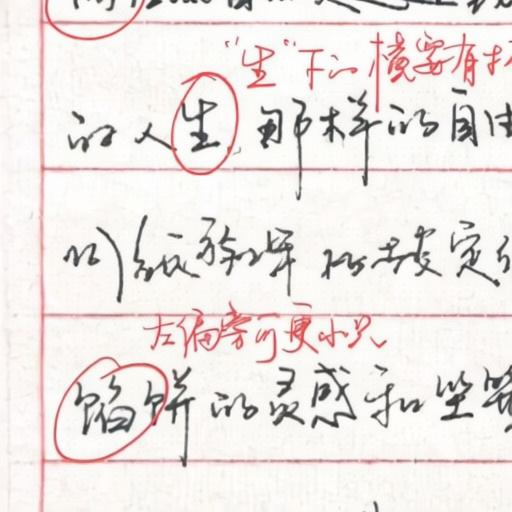} &
\includegraphics[width=0.11\textwidth]{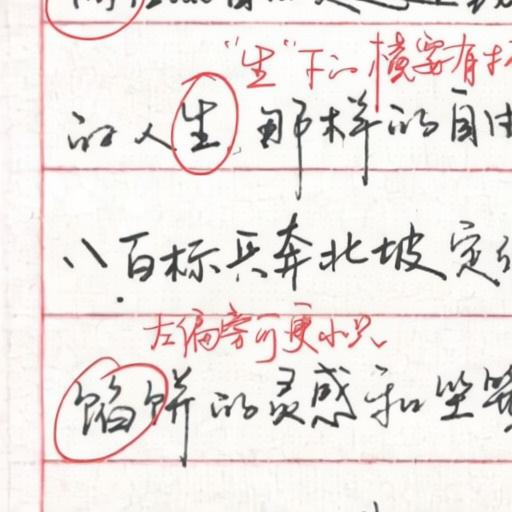} &
\includegraphics[width=0.11\textwidth]{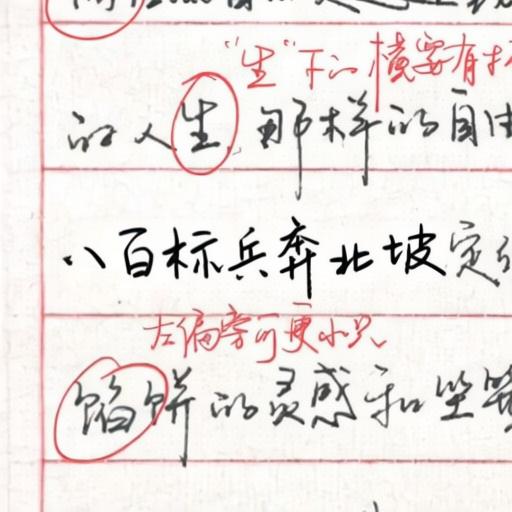} \\
original+mask & CFG=1 & CFG=3 & CFG=5 \\[6pt]
% Second row (single image)
\multicolumn{4}{c}{Prompt: "\includegraphics{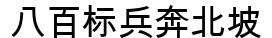}"} \\
\end{tabular}
\caption{Effect of classifier-free guidance (CFG). Original image with target area mask shown leftmost. Without CFG (CFG=1), TextMastero produces unreadable text. CFG=3 improves readability while maintaining style. CFG=5 generates overly thick text, deviating from the original region.}
\label{fig:cfg}
\end{figure}

\subsubsection{Classifier-Free Guidance}
Classifier-free guidance (CFG) has been shown to be effective in controlling the strength of prompt-following ability in text-to-image LDMs ~\cite{DBLP:conf/cvpr/RombachBLEO22}. Hence, we train our scene text editing model with a certain probability of null glyph condition to leverage the CFG strength. Our experiments with CFG reveal a crucial trade-off in scene text editing. As demonstrated in Figure \ref{fig:cfg}, we found that in inference, a higher CFG scale results in stronger glyph control, producing clearer and thicker text. This allows for improved readability when editing texts. However, our findings show that this comes at a cost to style preservation. Conversely, lower CFG scales excel at maintaining the original text style, though occasionally at the expense of target text accuracy. This insight offers a new approach to balancing readability and style preservation in scene text editing tasks.

\subsection{Latent Guidance Module}
The latent guidance module mainly controls the visual appearance of the generated text (i.e., style control). Figure \ref{fig:combined} (bottom) demonstrates our latent guidance module. The first three inputs, $x, m, x_m$, are the standard in-painting formulation proposed by ~\cite{DBLP:conf/cvpr/RombachBLEO22}, where $x$ and $x_m$ are encoded by VAE to become latent $4 \times H / 8 \times W / 8$, and the mask $m$ is interpolated to $1 \times H / 8 \times W / 8$. Together, they form a 9-channel input feature map for UNet $z^{inpaint}_t$. Then it is fed into the input convolution of the UNet to produce $y^{inpaint}_t$. In addition to that, we introduce a convolutional style encoder $E_{style}$ to capture the nuances of the style from the masked region $c_{style}$, as a time-independent latent feature $y^{glyph}$. Following ~\cite{DBLP:conf/iccv/ZhangRA23}, we use zero convolution improve representation quality. It reads:
\begin{equation}
    y^{style} = ZeroModule(E_{style}(c_{style}))
\end{equation}
Finally, we concatenate $y^{inpaint}_t$, and $y^{style}$ and use another convolution layer to match the UNet's original input dimensions. That is,
\begin{equation}
    z_t = Conv(Concat(y^{inpaint}_t, y^{style}))
\end{equation}
This design effectively encapsulates position information, glyph style, and style nuances at the same time.

\begin{figure*}[htbp]
\centering
\begin{tabular}{ccccc}
\emph{Prompt} & \emph{Masked Source Image} & {DiffUTE} & {AnyText} & \textbf{Ours} \\
\midrule
\raisebox{-.5\height}{"\includegraphics{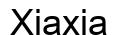}"} &
\raisebox{-.5\height}{\includegraphics[width=0.18\textwidth]{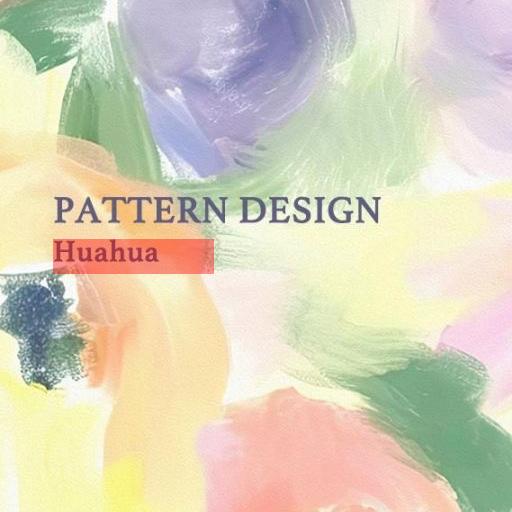}} &
\raisebox{-.5\height}{\includegraphics[width=0.18\textwidth]{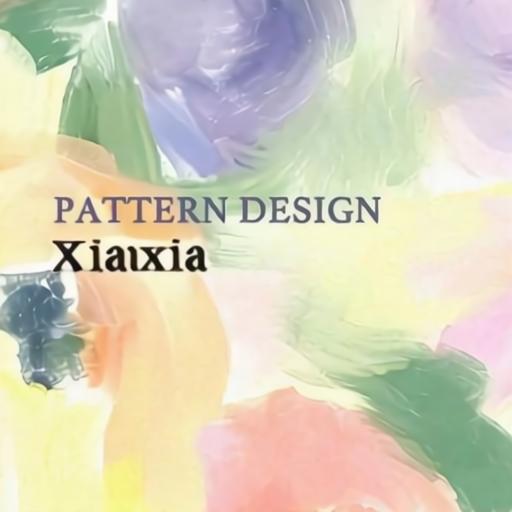}} &
\raisebox{-.5\height}{\includegraphics[width=0.18\textwidth]{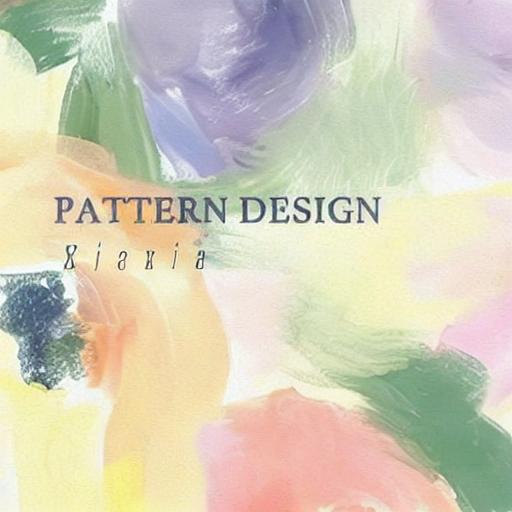}} &
\raisebox{-.5\height}{\includegraphics[width=0.18\textwidth]{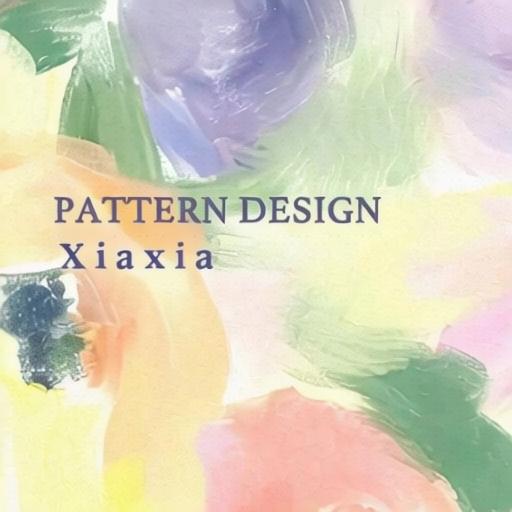}} \\
\raisebox{-.5\height}{"\includegraphics{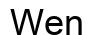}"} &
\raisebox{-.5\height}{\includegraphics[width=0.18\textwidth]{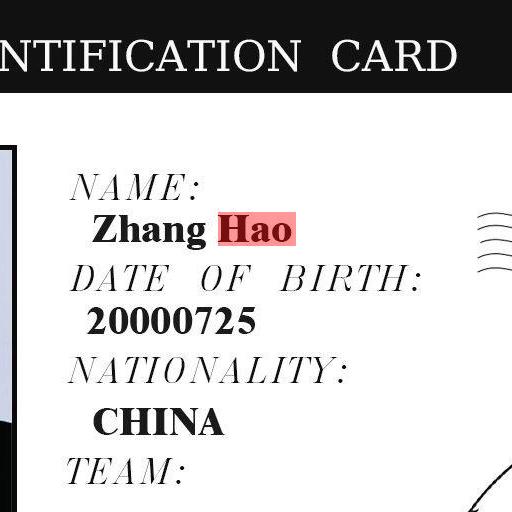}} &
\raisebox{-.5\height}{\includegraphics[width=0.18\textwidth]{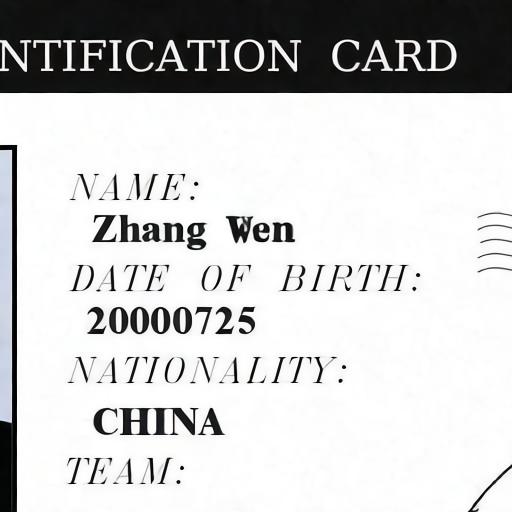}} &
\raisebox{-.5\height}{\includegraphics[width=0.18\textwidth]{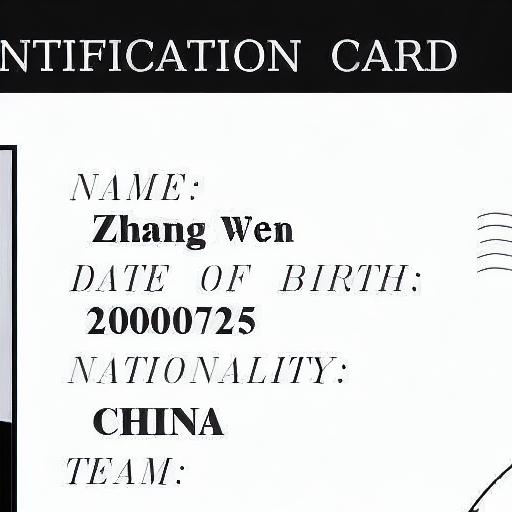}} &
\raisebox{-.5\height}{\includegraphics[width=0.18\textwidth]{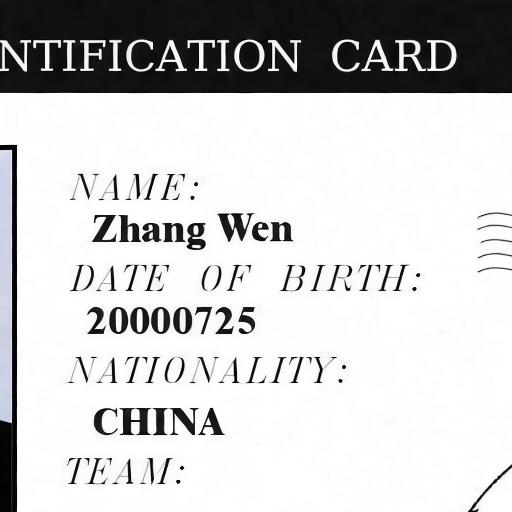}} \\
\raisebox{-.5\height}{"\includegraphics{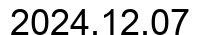}"} &
\raisebox{-.5\height}{\includegraphics[width=0.18\textwidth]{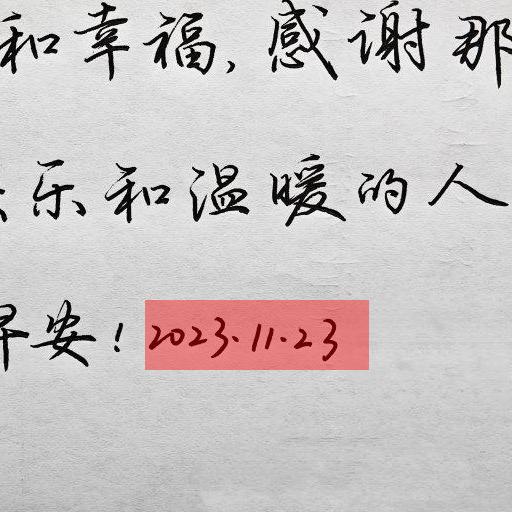}} &
\raisebox{-.5\height}{\includegraphics[width=0.18\textwidth]{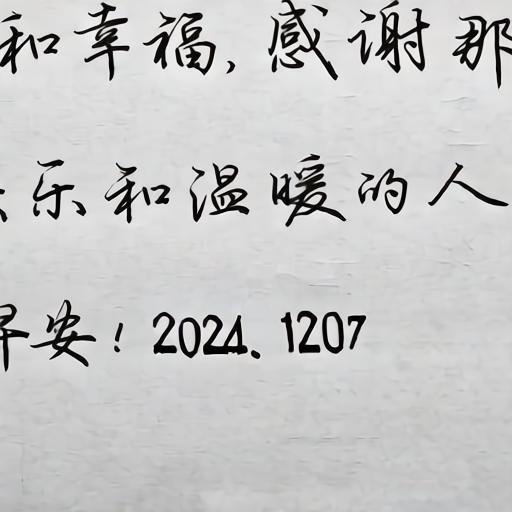}} &
\raisebox{-.5\height}{\includegraphics[width=0.18\textwidth]{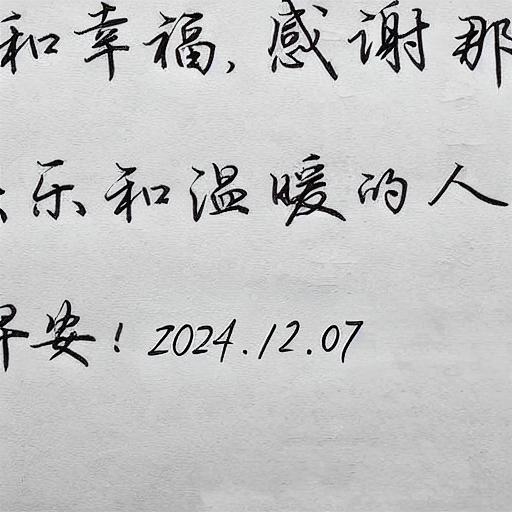}} &
\raisebox{-.5\height}{\includegraphics[width=0.18\textwidth]{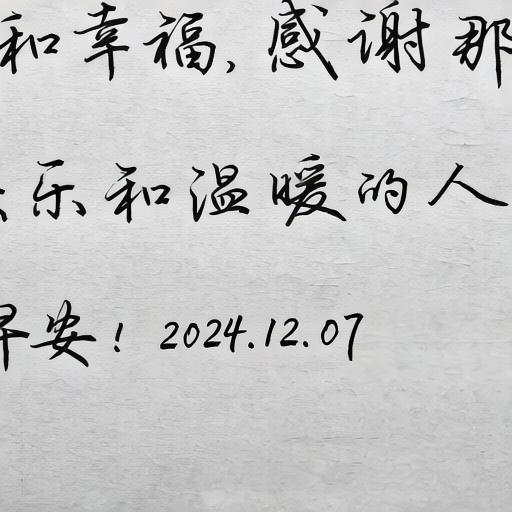}} \\
\raisebox{-.5\height}{"\includegraphics{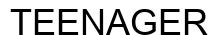}"} &
\raisebox{-.5\height}{\includegraphics[width=0.18\textwidth]{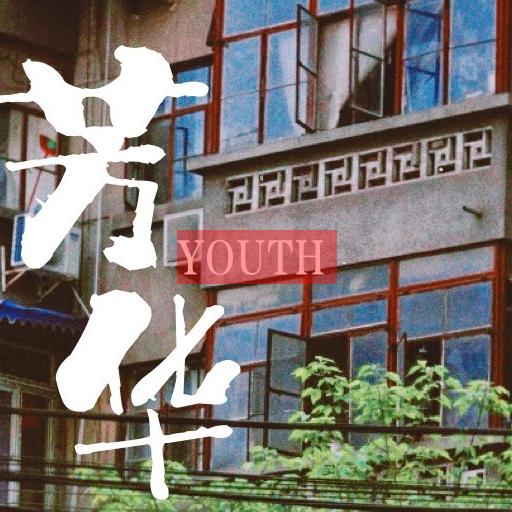}} &
\raisebox{-.5\height}{\includegraphics[width=0.18\textwidth]{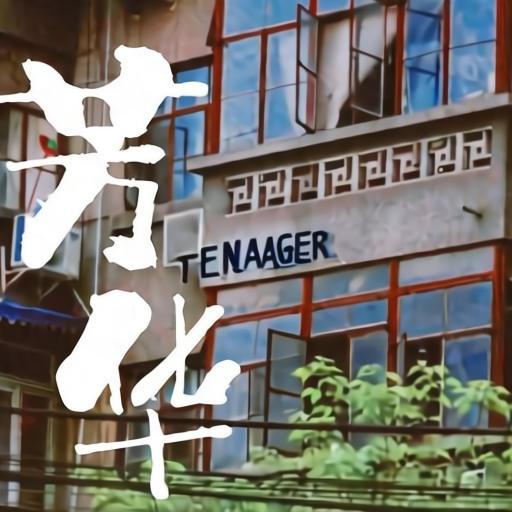}} &
\raisebox{-.5\height}{\includegraphics[width=0.18\textwidth]{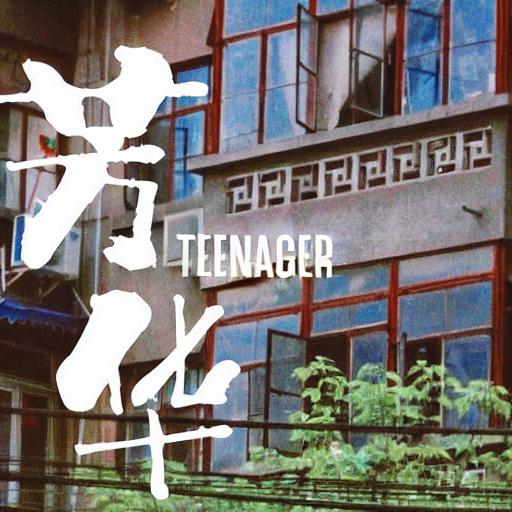}} &
\raisebox{-.5\height}{\includegraphics[width=0.18\textwidth]{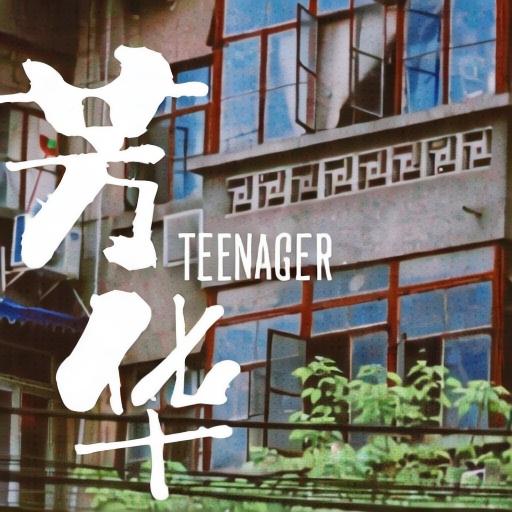}} \\
\raisebox{-.5\height}{"\includegraphics{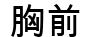}"} &
\raisebox{-.5\height}{\includegraphics[width=0.18\textwidth]{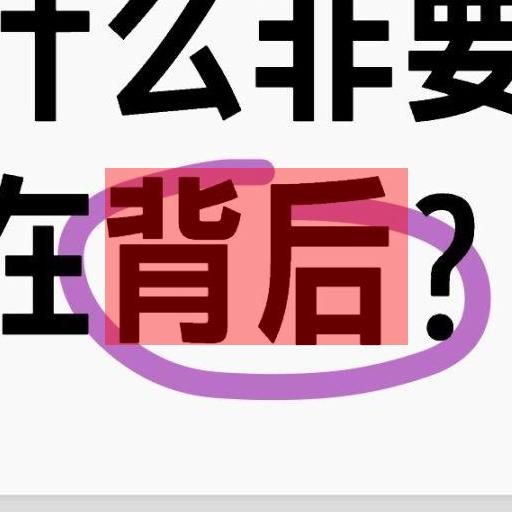}} &
\raisebox{-.5\height}{\includegraphics[width=0.18\textwidth]{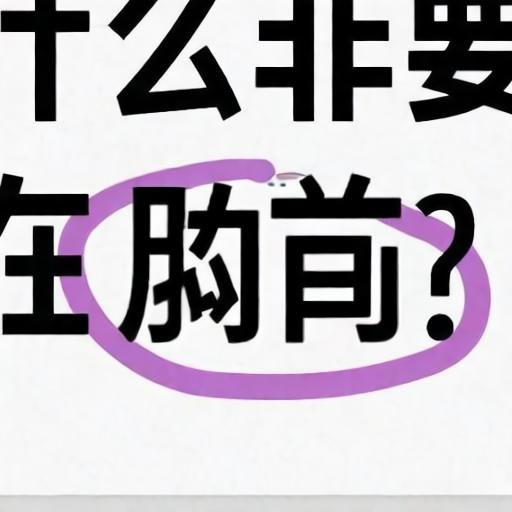}} &
\raisebox{-.5\height}{\includegraphics[width=0.18\textwidth]{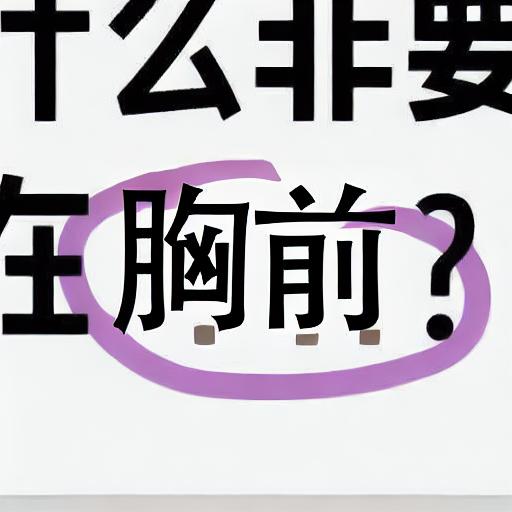}} &
\raisebox{-.5\height}{\includegraphics[width=0.18\textwidth]{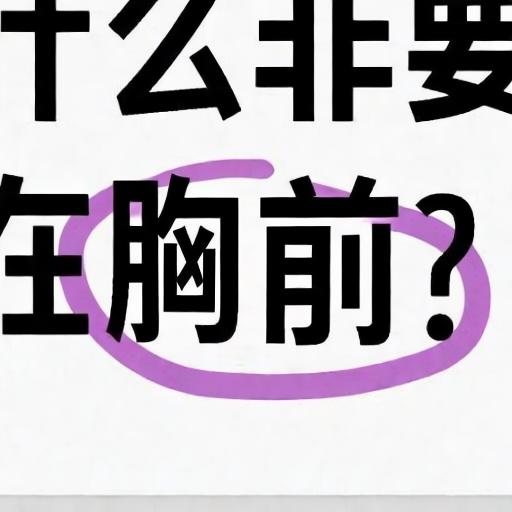}} \\
\raisebox{-.5\height}{"\includegraphics{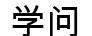}"} &
\raisebox{-.5\height}{\includegraphics[width=0.18\textwidth]{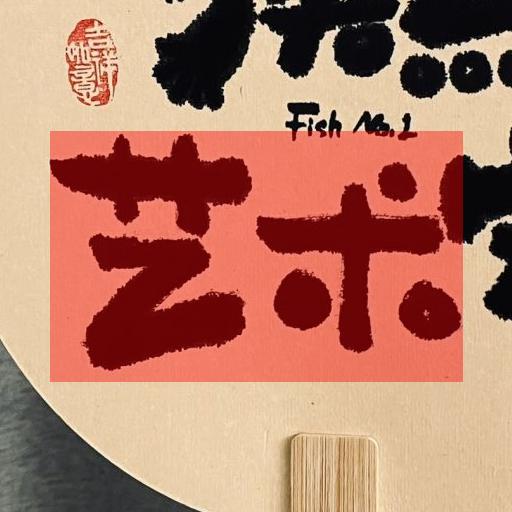}} &
\raisebox{-.5\height}{\includegraphics[width=0.18\textwidth]{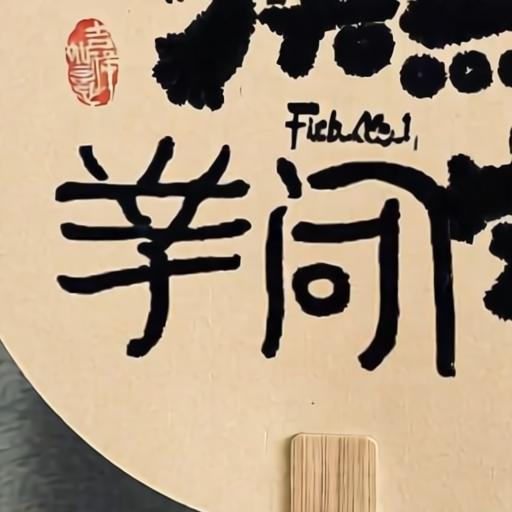}} &
\raisebox{-.5\height}{\includegraphics[width=0.18\textwidth]{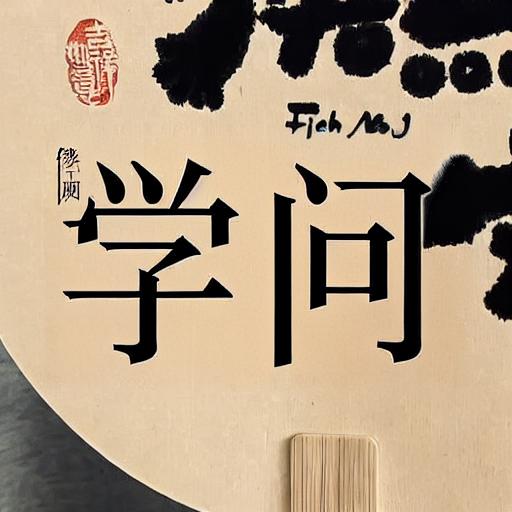}} &
\raisebox{-.5\height}{\includegraphics[width=0.18\textwidth]{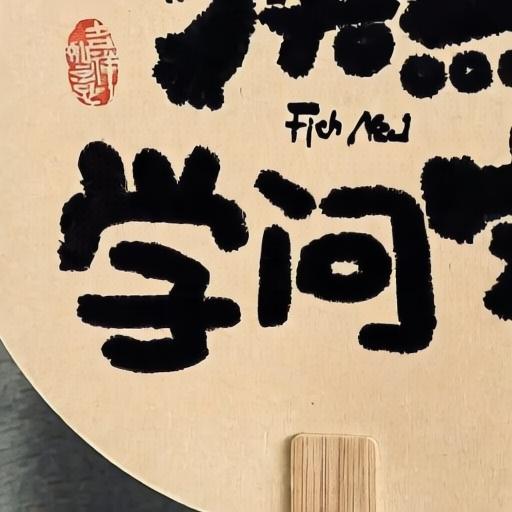}} \\
\raisebox{-.5\height}{"\includegraphics{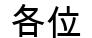}"} &
\raisebox{-.5\height}{\includegraphics[width=0.18\textwidth]{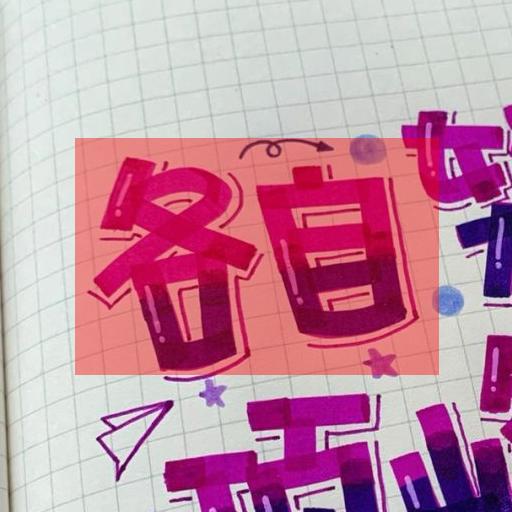}} &
\raisebox{-.5\height}{\includegraphics[width=0.18\textwidth]{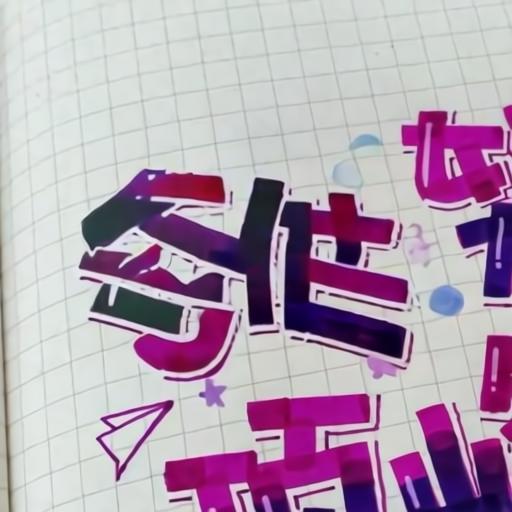}} &
\raisebox{-.5\height}{\includegraphics[width=0.18\textwidth]{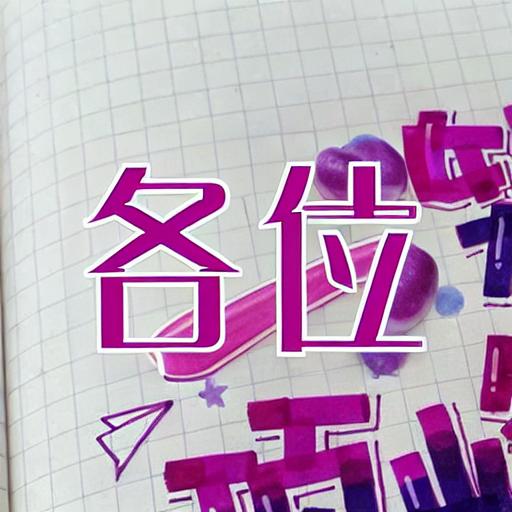}} &
\raisebox{-.5\height}{\includegraphics[width=0.18\textwidth]{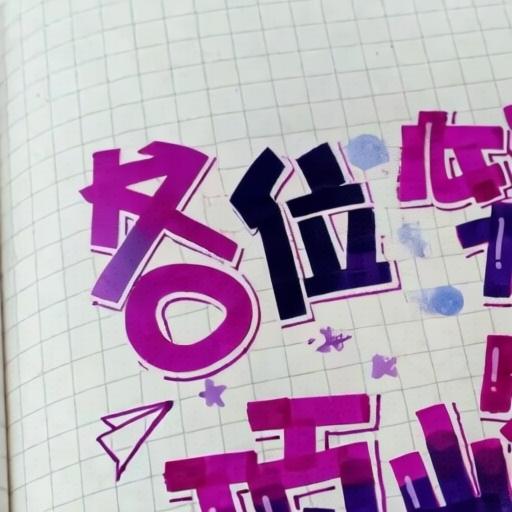}} \\
\end{tabular}
\caption{Comparison of scent text editing methods: DiffUTE, AnyText, and our \emph{TextMastero}. More results are available in supplementary materials.}
\label{fig:method_comparison}
\end{figure*}

\begin{table*}[t]
\centering
\begin{tabular}{l cccc cccc}
\multirow{3}{*}{Model} & \multicolumn{4}{c}{Text Accuracy} & \multicolumn{4}{c}{Style Similarity} \\
 & \multicolumn{2}{c}{Sen.Acc $\uparrow$} & \multicolumn{2}{c}{CER $\downarrow$} & 
 \multicolumn{2}{c}{FID $\downarrow$} & \multicolumn{2}{c}{Avg.LPIPS $\downarrow$} \\
 & English & Chinese & English & Chinese & English & Chinese & English & Chinese \\
DiffUTE & 0.3319 & 0.2523 & 0.3186 & 0.4048 & 14.3176 & 24.9295 & 0.1313 & 0.2056 \\
AnyText & 0.6067 & 0.5801 & 0.1730 & 0.2088 & 10.4257 & 24.9004 & 0.1098 & 0.1978 \\
Ours    & \textbf{0.8170} & \textbf{0.7301} & \textbf{0.0741} & \textbf{0.1341} & \textbf{4.6101} & \textbf{11.8915} & \textbf{0.0545} & \textbf{0.1007} \\
\end{tabular}
\caption{Model comparison results.}
\label{tab:model_comparison}
\end{table*}

\section{Experiments}

\subsection{Dataset}
We use AnyWord-3M~\cite{DBLP:conf/iclr/TuoXHGX24}, a dataset for scene text generation and scene text editing, as our training set. We utilize the dataset specifically for scene text editing scanario: each sample is an image containing multiple line-level annotations, which consists of text content, and its corresponding polygonal position as target area to generate new text. Since each image can have multiple text regions, we randomly sample one text-position pair in training. Our dataset formulation can be written as:
$$\mathcal{D} = \{(I_i, \{(T_{ij}, P_{ij})\}_{j=1}^{M_i})\}_{i=1}^N$$
$N$ is the total number of images, approximately 3.5 million in case of AnyWord dataset. $M$ is number of text-position pairs in the $i$th image. Hence $(T_{ij}, P_{ij})$ would be a single data sample. 

\subsection{Implementation Details} 
We initialize our VAE and UNet with stable-diffusion 2.1 weights~\cite{DBLP:conf/cvpr/RombachBLEO22}. Each glyph Transformer has four layers and two heads, and $d_{output} = 1024$ to match with UNet's conditional embedding dimension; their weights are initialized randomly. Latent guidance module's style encoder is a two-layer convolutional module mapping style image $c_{style} \in \mathbb{R}^{3 \times 128 \times 128}$ to $y_{style} \in \mathbb{R}^{128 \times 64 \times 64}$, to allow concatenation with $y^{inpaint}_t \in \mathbb{R}^{320 \times 64 \times 64}$, resulting $\mathbb{R}^{448 \times 64 \times 64}$ intermediate latents. Finally, a zero-convolution layer is used to map $\mathbb{R}^{448 \times 64 \times 64}$ back to $\mathbb{R}^{320 \times 64 \times 64}$. That gives us the final style-guided latent $z_t$. Cross attention condition $e_{final}$ and latent guidance $z_t$ are then used to train the UNet. Our model was trained for 15 epochs with a global batch size of 256 on 8 V100S-32G cards. We also use a null condition probability of 0.1 to allow classifier free guidance in inference.

\subsection{Comparisons}

\subsubsection{Setup}
We evaluate our model using AnyText-Eval~\cite{DBLP:conf/iclr/TuoXHGX24} for quantitative analysis (2000 images, 4181 English and 2092 Chinese text-position pairs) and a curated dataset of 80 challenging stylistic images (120 text-position pairs) for qualitative assessment. While AnyText-Eval provides statistical coverage, its target texts match the original, not fully reflecting real-world editing tasks. Our curated dataset focuses on challenging stylistic variations and incorporates new text contents. 

We benchmark against state-of-the-art models DiffUTE~\cite{DBLP:conf/nips/ChenXGLZLMZW23} and AnyText. For AnyText, we use the public AnyWord-3M checkpoint. We trained DiffUTE on AnyWord-3M for 15 epochs with a batch size of 256 and null condition probability of 0.1, matching our model's hyperparameters. All methods use DDIM~\cite{DBLP:conf/iclr/SongME21} sampler with 20 steps denoising steps during inference, with CFG scale 9 for AnyText and 3 for ours and DiffUTE. 

While other diffusion-based scene text editing methods exist (e.g., TextDiffuser~\cite{DBLP:conf/nips/ChenHL0CW23}, GlyphDraw~\cite{DBLP:journals/corr/abs-2303-17870}, and GlyphControl~\cite{DBLP:conf/nips/YangGYLDH023}), we focus on DiffUTE and AnyText for two reasons: they are multilingual, aligning with our emphasis on multilingual scene text editing, and AnyText has demonstrated significant superiority over these alternatives in comprehensive evaluations. This approach ensures our comparison is against the most advanced methods in the field.

\begin{table*}[htbp]
\centering
\setlength{\tabcolsep}{3pt}  % Reduced to accommodate all columns
\renewcommand{\arraystretch}{1.2}
\begin{tabular}{l cccc cccc}
\multirow{3}{*}{Ablation Setting} & \multicolumn{4}{c}{Text Accuracy} & \multicolumn{4}{c}{Style Similarity} \\
 & \multicolumn{2}{c}{Sen.Acc $\uparrow$} & \multicolumn{2}{c}{CER $\downarrow$} & \multicolumn{2}{c}{FID $\downarrow$} & \multicolumn{2}{c}{Avg.LPIPS $\downarrow$} \\
 & English & Chinese & English & Chinese & English & Chinese & English & Chinese \\
\multicolumn{9}{l}{\textit{Glyph Conditioning Module}} \\
Full model & \textbf{0.5494} & \textbf{0.5120} & \textbf{0.1766} & \textbf{0.2367} & {30.9095} & {51.3762} & \textbf{0.1190} & {0.2134} \\
w/o multi-level fusion & 0.4536 & 0.3698 & 0.2470 & 0.3314 & 32.4550 & 53.0127 & 0.1247 & 0.2208 \\
w/o backbone feature (1 glyph TFM) & 0.5065 & 0.4271 & 0.2128 & 0.2987 & \textbf{30.5866} & \textbf{49.6588} & {0.1196} & \textbf{0.2110} \\
w/o neck cross attention (0 glyph TFM) & 0.3263 & 0.2735 & 0.3137 & 0.3916 & 30.6271 & 51.6288 & 0.1211 & 0.2121 \\
\multicolumn{9}{l}{\textit{Latent Guidance Module}} \\
Full model & \textbf{0.5494} & \textbf{0.5120} & \textbf{0.1766} & \textbf{0.2367} & \textbf{30.9095} & \textbf{51.3762} & \textbf{0.1190} & \textbf{0.2134} \\
w/o style encoder & 0.5150 & 0.5015 & 0.1925 & 0.2465 & 35.8388 & 53.0539 & 0.1405 & 0.2229 \\
\end{tabular}
\caption{Ablation study results.}
\label{tab:ablation_results}
\end{table*}

\subsubsection{Quantitative} 
We measure four metrics for quantitative studies: sentence accuracy (Sen.Acc) and character error rate (CER) for text content fidelity; Fréchet Inception Distance (FID)\cite{Seitzer2020FID} and LPIPS~\cite{zhang2018perceptual} for style similarity. Sen.Acc measures line-level accuracy, while CER is for character-level accuracy. FID measures distribution-level style similarity, while LPIPS focuses on sample-level similarity. We average LPIPS distances over all samples for the final measurement. FID and LPIPS are measured between cropped ground truth image and generated text regions, as scene text editing aims to maintain similarity in the edited region while leaving non-target regions unchanged.

Table \ref{tab:model_comparison} shows our method significantly outperforms prior arts in both text generation accuracy and style similarity. Our overall sentence accuracy (averaging English and Chinese results) is 48.14\% and 18.02\% higher than DiffUTE and AnyText, respectively; CER is 25.76\% and 8.68\% lower. Note that the evaluation OCR predictor ~\cite{modelscope2023duguangocr} often fails to recognize stylish characters our model generates correctly, suggesting actual accuracy may be higher than reported. Our method also better maintains style similarity, with substantially lower FID and LPIPS distances compared to DiffUTE and AnyText.

\subsubsection{Qualitative}
Figure \ref{fig:method_comparison} shows some comparison results among our method, AnyText, and DiffUTE. Our method is substantially better than prior arts on both text accuracy and style similarity. More qualitative results are available in the supplementary materials.

\subsection{Ablations}
We evaluate each component's effectiveness by ablating them sequentially. For the glyph conditioning module, we first remove multi-level fusion, then backbone features entirely (eliminating one glyph transformer), and finally both glyph transformers, resulting in a vanilla structure using only character-level glyph images to extract $N \times 720$ features. We add absolute positional embeddings and project them to $N \times 1024$ to be compatible with SD2.1's UNet. For the latent guidance module, we ablate our style encoder. All experiments used a 375K image subset of Anyword-3M, trained for 15 epochs with a batch size of 256 on 4 A100-40G cards.

Table \ref{tab:ablation_results} shows quantitative results. Without multi-level fusion, we see an 11.9\% average sentence accuracy drop for English and Chinese texts. Removing backbone features and one glyph transformer results in a 6.39\% drop, performing better than using only the last hidden states of OCR backbone. This suggests that backbone and neck features have distinct distributions, which our FPN-like fusion module addresses by introducing parameters for multi-level feature fusion. The vanilla module shows a 23.08\% sentence accuracy drop. CER results verify these trends.

Style similarity remains consistent across glyph conditioning ablations, as expected since these primarily affect text accuracy. For the latent guidance module, average FID for English and Chinese generations increases by 3.3035, and average Avg.LPIPS by 0.0845, indicating our style encoder's role in maintaining style similarity post-editing.

\section{Conclusion \& Future Works}
We introduced TextMastero, a novel framework for multilingual scene text editing based on latent diffusion models. Our approach addresses challenges in generating complex glyphs, particularly CJK characters, in stylistic scenes. Key innovations include a glyph conditioning module for fine-grained content control and a latent guidance module for style consistency. Experiments demonstrate TextMastero's superior performance in text fidelity and style similarity across diverse languages and visual scenarios. Future work could enhance style similarity and generation stability using networks like VGG~\cite{DBLP:journals/corr/SimonyanZ14a} for style feature extraction, and expand support to more minority languages beyond CJK to increase impact and accessibility.

%%%%%%%%%%%%%%%%%%%% END OF EXP %%%%%%%%%%%%%%%%

\bibliographystyle{unsrtnat}
\bibliography{ms}  %%% Uncomment this line and comment out the ``thebibliography'' section below to use the external .bib file (using bibtex) .




% 	\bibitem{keshet2016prediction}
% 	Keshet, Renato, Alina Maor, and George Kour.
% 	\newblock Prediction-Based, Prioritized Market-Share Insight Extraction.
% 	\newblock In {\em Advanced Data Mining and Applications (ADMA), 2016 12th International 
%                       Conference of}, pages 81--94,2016.

% \end{thebibliography}

\end{document}